# Reproduction of Images by Gamut Mapping and Creation of New Test Charts in Prepress Process

Jaswinder Singh Dilawari, Dr. Ravinder Khanna

**ABSTARCT**

With the advent of digital images the problem of keeping picture visualization uniformity arises because each printing or scanning device has its own color chart. So, universal color profiles are made by ICC to bring uniformity in various types of devices. Keeping that color profile in mind various new color charts are created and calibrated with the help of standard IT8 test charts available in the market. The main objective to color reproduction is to produce the identical picture at device output. For that principles for gamut mapping has been designed.

—————— ◆ ——————

## 1. INTRODUCTION

When an image is printed using different devices or scanners then problem of rendering of colors arises. This can be clearly pictured in figure 1(© FujiFilm 2002)

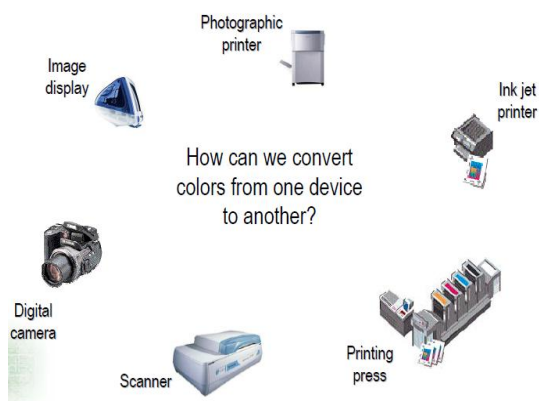

Figure 1: Each device sees color differently

To avoid this common profile by ICC has been made. ICC profile is a data file describing the color characteristics of an imaging device [1]. Figure 2 visualizes the problem in printing of image from different devices.

- Jaswinder Singh Dilawari is a Ph.D Research Scholar at Pacific University,Udaipur,Rajasthan (India), Ph-9896348400, E-mail- dilawari.jaswinder@gmail.com
- Dr. Ravinder Khanna is currently working as Principal in Sachdeva Engineering College for Girls, Mohali, Punjab (India). Ph- 7814930123,E-mail -ravikh_2006@yahoo.com

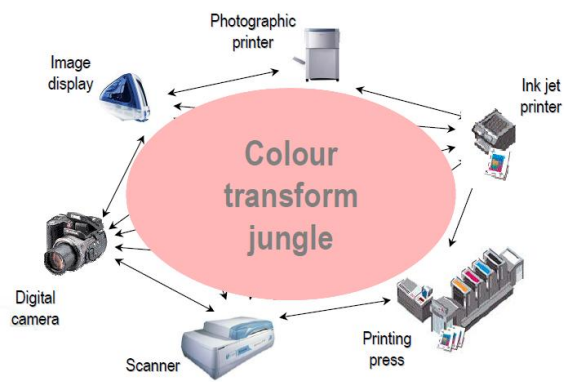

Figure 2: Color rendering problem due to various devices

The primary purpose of use of this file is to maintain color consistency in images viewed, displayed or printed on various devices. ICC profile mainly converts images form RGB to CMYK because digital images used by prepress department are in RGB color format that has to be converted to CMYK color format before printing [2]. Figure 3 shows the working of ICC profile. The profiles contain information about separation, black start, black width, total ink coverage. A device is characterized by the color chart used in that device. By using the common format for characterization of color units it is easier to determine the color gamut of device and to optimize the image to produce high image quality. Various color gamut are available in the market which perform well for all type of images whether that is low key tone image, high tone image or middle tone image [3]. Various methods had been adopted for producing color charts so that image quality in printing can be improved.

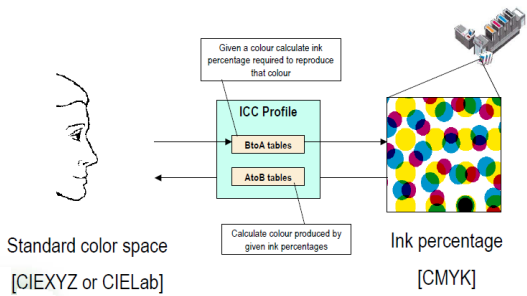

Figure3: Working of ICC profile

According to a paper published in TAGA 2003, Montreal, Canada a new color chart has been developed based on technical and visual image category analysis. This paper describes that image categorization using the adapted color charts improves the analysis of relevant image information with regard to both image gradation and detail reproduction. This new color chart preserves the image information even in low key areas and gives the better image fidelity to the original image [4]. Each scanner holds different color gamut. IT-8 test charts are used to correlate these scanners [5]. Color separation is used to control the amount of black, cyan, magenta and yellow needed to produce the different tones in conversion from RGB to CMYK. GCR (Gray Component Replacement) and UCR (Under Color Removal) are the two main color separation techniques. These separation techniques can be optimized for different paper stocks in order to achieve a good tone distribution. The total amount of ink used in a printing process must normally be reduced in order to avoid printing problems such as slurring and quality problems such as lack of image detail [6]. To combine the gray scale image in the print a novel spectral imaging technique is used. Just like conventional color imaging, IMI exploits the interaction between the illuminating light and colorants used for printing and the manner in which the eye adapts to illumination [7][8]. Selection of color gamut also depends upon the display device because it color gamut then decide the power consumption, resolution and viewing angle of that display device.

## 2. CREATION OF NEW TEST CHARTS

New test charts used for output characterization of output are created by evaluating tone steps. A new set of color values was used to create an image-adapted test chart, different from the gamma and gradation values normally used. Spectral measurements were made on the new test charts, and new output profiles were calculated and applied in the RGB-to-CMYK conversion for the specific image category aimed for. The advantage of producing your own test chart is that it is possible to achieve a better match to the originals being scanned. A validation print was made with the new separation values applied to the specific image category aimed for. The new image adapted color chart is created by judging the borders of the L* values for the different images of different categories. Different images used to judge L* values border are shown in figure 4.

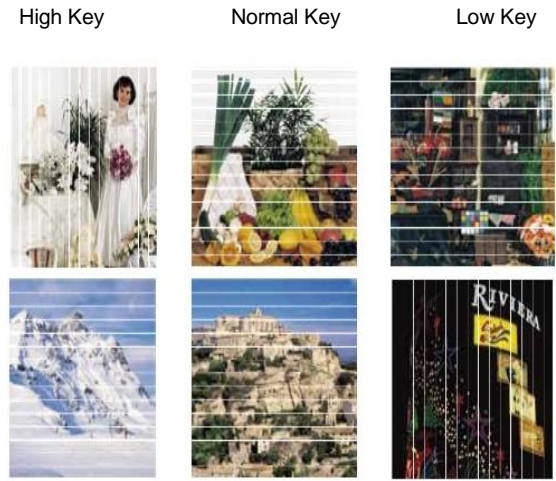

Figure 4: Different types of images used to find L* values

The distribution of the L*-value in the three types of images indicates the borders that may exist between these images, Figure 5

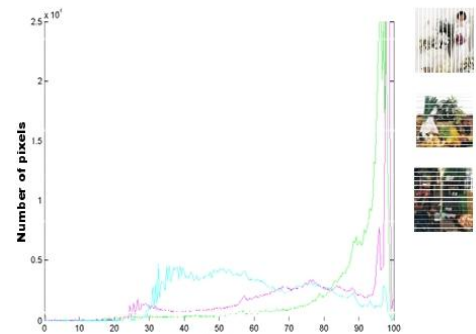

Figure 5: Distribution of L*-value for high-key, normal-key and low-key images. The peak for high L-values in the normal-key image is caused by the white background. The graph was created in Matlab (Enoksson, 2001).

These images are further processed in matlab and their histograms are studied as shown in figure 6

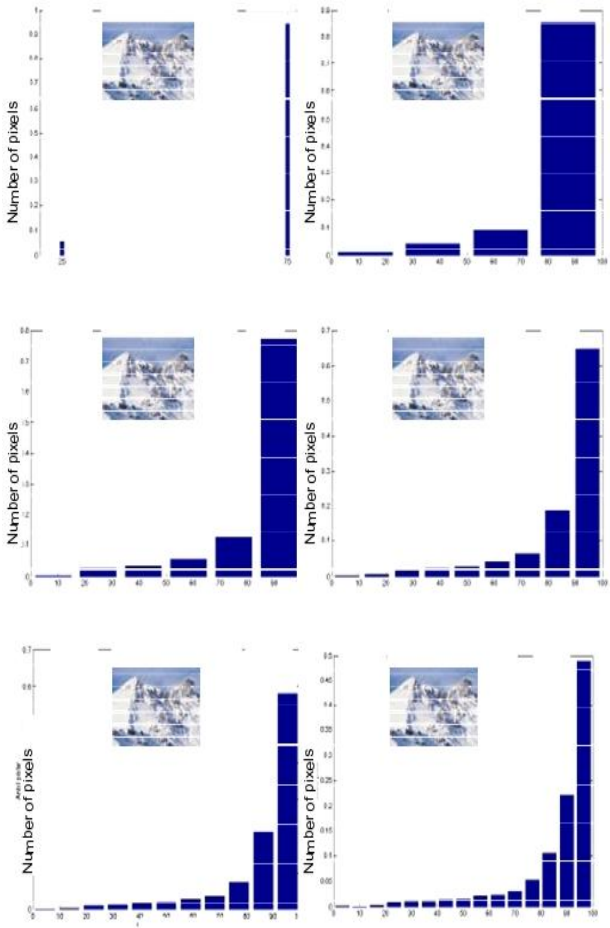

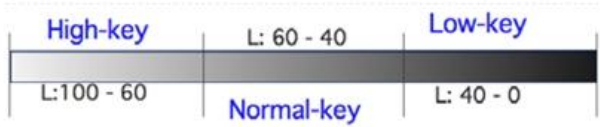

Figure 6: The steps made in the Matlab analysis. Number of pixels in different steps of L* scale in high-key image. The steps made it possible to find the borders between the images

The studies of the three image categories (high-key, low-key, normal-key) revealed that the borders in the L*-scale for high-key images were 100-60, for normal-key images 60-40, and for low-key images 40-0, Figure 7 [9]

Figure 7: Border range of L* values for different images

This border information is used to develop new adapted color chart. An example of new image color chart developed by above L* values is shown in figure 8.

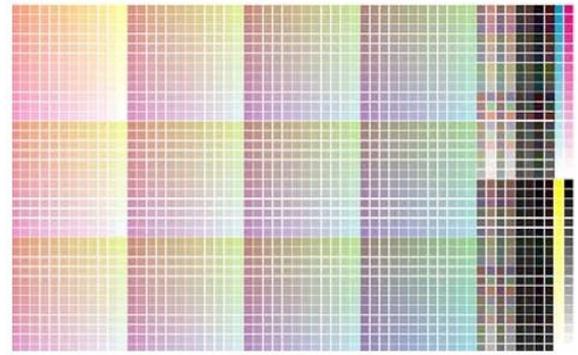

Figure 8: An example of the new image-adapted test chart

These developed color charts are than compared with available color charts in the market and different tonal values from figure 6 is used for that purpose [10].

## 3. IT-8 TEST CHARTS

The analysis of borders of L* values gives rise to the adoption of IT-8 Test charts. There are several vendors producing IT8-targets for scanner characterization [11]. The targets follow a certain pattern based on ISO standardization values in LCH (ISO 12641-1997).

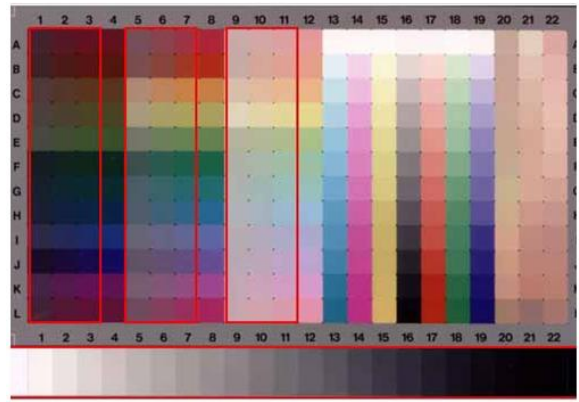

Figure 9: The scanner target consists of a total of 264 colors. The red frames show the standardized values.

Twelve separate hue angles are defined at three separate lightness levels. For each specific hue angle and luminance level, there are four different chroma values. The highest chroma value is defined as the maximum chroma which can be generated on a given medium with no change in the hue angle and lightness level. A further 84 patches provide additional tone scales which are not defined by any ISO-standard. Seven tone scales are defined for the colors cyan, magenta, yellow, red, green and blue (no ISO standard defined). Each tone scale is built-up in twelve steps starting from the lowest chroma value and keeping the hue angle stable. Each vendor has defined an optimal tone scale for their own specific output media.

The last three columns in the test chart are vendor-specific. Here the vendor manufacturing a target was allowed to add any feature they deemed worthwhile. Each vendor has chosen to use this area differently[12][13]. Kodak has chosen the image of a model and several skin tone patches; Agfa and Fuji have both chosen to have patches of special colors in this area (McDowell, 2002).

The color charts in a color gamut varies even using a single setting. So to capture an image more accurately with a certain color gamut, scanner color gamut has to be large enough so that image color gamut falls in that. So reference reading of IT8 test chart is must to customize own color chart to improve the scanning picture quality. In customized color charts, customized color patches can be added for specific color values as shown in figure 10.

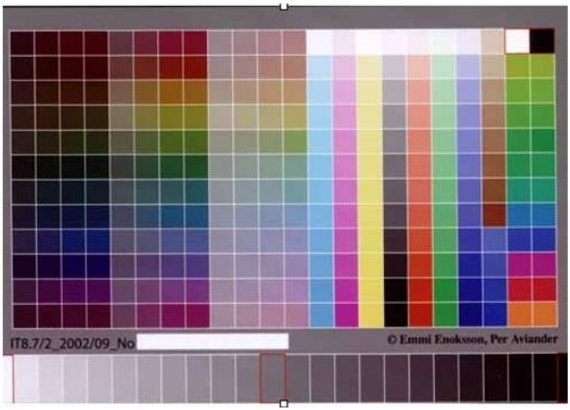

Figure 10: The customized IT.8 target for scanners

## 4. GAMUT MAPPING

The objective in color reproduction is to produce an identical color appearance. Doing so does not usually mean producing an exact match of tristimulus values as color appearance is affected by many other variables, such as the state of adaptation or the overall viewing conditions. Any mapping that solves the problem must relate three gamuts: that of the source device, that of the destination device, and that of the image. Each device has own color gamut so a gamut transformation must takes place before image reproduction. The gamut transformation is done such that transformation only changes the tristimulus value preserving the appearance of the image. To accomplish this, the transformations are based on widely accepted graphic arts and psychophysical principles, which are the most precise definition available of what it means to "preserve the appearance" of an image. We summarize the principles as follows:

(1) The gray axis of the image should be preserved.
(2) Maximum luminance contrast is desirable.
(3) Few colors should lie outside the destination gamut.
(4) Hue and saturation shifts should be minimized.
(5) It is better to increase than to decrease the color saturation [14].

The transformation of gamut keeping above rules in mind should begin with elementary transformation. These transformations are combined and adjusted until the image gamut fits properly into the destination gamut. The combination and adjustment, which must include human aesthetic judgment, are performed interactively. To judge this visualization tools are required. The suggested tools are:

(1) Tools for true 3-D visualization of the problem areas in the gamuts, including interpenetrating, solid, shaded 3-D models of the image and destination gamuts. Our experience suggests that useful 3-D visualization tools need sophisticated lighting controls, and a precise mechanism for orienting and moving light sources and the viewpoint. The importance of concavities in gamut surfaces suggests the use of shadows or of curvature encoding techniques like those reported by Forrest [15]. Careful false-coloring of the gamut surfaces may help to orient the observer. Transparency would be useful for simultaneous viewing of the source and destination gamuts which necessarily intersect.

(2) An interactive tool to indicate source image pixels corresponding to a selected color in a gamut disp1a.y and its converse.

(3) Automatic and simultaneous highlighting of all out-of-gamut colors, both in gamut displays and in the source image.

(4) Interactive feedback showing changes in a gamut display when transformation parameters are varied.
(5) Interactive tools for defining gamut mapping transformations. Specifying translation distances, scaling, and rotation factors by pointing at positions in gamut plots would be very useful [7] [8].

These tools provide visualization whether image gamut fits properly into the destination gamut.

## 5. CONCLUSION

To avoid color rendering in different devices new color charts are created and gamut mapping is done. Creation of new color charts depend upon the border of L* values. To define a sharp boundary for low key image, high key image and middle key image based on L* values is almost impossible. So color charts created although work better for low key images but not for high key images so effectively. Gamut mapping found to be more effective because transformation of gamuts starts from elementary transformation and principles of human visual system are taken into consideration.